\documentclass[conference]{IEEEtran}

\usepackage{graphicx}
\usepackage{caption}
\usepackage{subcaption}
\usepackage{multirow}
\usepackage{tabularx}
\usepackage{float}
\usepackage{tipa}
\usepackage{hyperref}
\usepackage{xcolor}
\usepackage{todonotes}
\usepackage[utf8]{inputenc}
\usepackage{listings}
\usepackage{algorithm}
\usepackage{algpseudocode}
\usepackage{amsmath}
\usepackage{tikz}
\usepackage{pgfplots}
\usepackage{siunitx}
\usepackage{tabularx}
\usepackage{booktabs}
\usepackage{todonotes}
\usepackage{tabularray}
\usepackage{multirow}
\usepackage{subcaption}
\usepackage{amsfonts}
\usepackage{microtype}
\usepackage{makecell}

\definecolor{codegray}{rgb}{0.5,0.5,0.5}
\definecolor{backcolour}{rgb}{0.98,0.98,0.98}

\graphicspath{{figures/}}

\title{Cross-Domain Few-Shot Writer Adaptation for Real-World Handwritten Mathematical Expression Recognition}

\author{
    \IEEEauthorblockN{1\textsuperscript{st} Paulo Grane Gabriel Silva}
    \IEEEauthorblockA{\textit{Department of Computer Technology} \\
    \textit{De~La~Salle~University}\\
    Manila, Philippines \\
    paulo\_silva@dlsu.edu.ph}
    \and
    \IEEEauthorblockN{2\textsuperscript{nd} Lorenz Bernard Marqueses}
    \IEEEauthorblockA{\textit{Department of Computer Technology} \\
    \textit{De~La~Salle~University}\\
    Manila, Philippines \\
    lorenz\_marqueses@dlsu.edu.ph}
    \and
    \IEEEauthorblockN{3\textsuperscript{rd} Joel Ilao}
    \IEEEauthorblockA{\textit{Department of Computer Technology} \\
    \textit{De~La~Salle~University}\\
    Manila, Philippines \\
    joel.ilao@dlsu.edu.ph}
}

\begin{document}

\maketitle

\begin{abstract}
    Handwritten mathematical expression recognition (HMER) refers to the task of recognizing and converting handwritten mathematics into a parsable markup language, usually LaTeX. No current state-of-the-art-competitive system adjusts to the way a specific person writes, and the domain gap between training images (usually digital or perfectly binarized) and images physically taken with a camera used in inference has received fairly little attention for this specific problem. We characterize this domain gap through fragmentation and stroke-width analyses of the images as well as introduce a writer-adaptive fine-tuning pipeline to MFH-CoMER in an attempt to address it. We further introduce sample author-specific datasets, consisting of five handwriting category subsets from two authors, and evaluate using a McNemar's test and permutation tests adapted to limited data. Results suggest an increase in expression recognition rate and a decrease in CER for digital handwriting but more varied results for physical handwriting, with the model struggling for handwriting articles that the base model can already evaluate well. Nonetheless, adapted models were found to have improved results for four out of five subsets. Statistical testing results suggest a consistency in improvement for two of the tested author-specific subsets. Our results point towards the potential feasibility of writer adaptation for the HMER task.
\end{abstract}


\section{Introduction}
Handwritten mathematical expression recognition (HMER) refers to the task of recognizing and converting visual mathematics to markup, mainly in LaTeX format. This task is further classified into offline and online HMER, with the latter considering the recorded trajectories of each stroke made by a digital pen on the surface of a device. This project mainly refers to offline HMER. While HMER is currently dominated by deep learning methods that have brought significant increase in expression recognition rate (ExpRate), particularly with attention-based \cite{fink_icdar_2023} and more specifically Transformer models \cite{yin_tst_2026, li_uni-mumer_2025}, visual confusion caused by ambiguously written expressions \cite{kawakatsu_gryphone_2026} and visually similar characters \cite{li_uni-mumer_2025, yin_return_2026} remain among the common underlying causes of errors for a lot of modern architectures. Among recent techniques aiming to reduce this confusion, there is no existing writer-adaptive approach that adjusts to and ``learns" from the specific handwriting of an individual author.
The primary objective of this work is therefore to develop a writer-adaptive hybrid system that uses computer vision to recognize handwritten mathematical expressions and convert them into their equivalent LaTeX code. This is to test the potential feasibility of introducing writer adaptation to the task of HMER to improve writer-specific evaluation. The ideal model adapts to the handwriting of an individual so that the generated LaTeX output is more accurate than what current writer independent methods are capable of making. Specifically, we implement a writer-dependent adaptation of the base MFH-CoMER model to a single writer using a few of their own samples. We then compare the adapted model against the baseline using expression recognition rate and character error rate which measures how much the writer adaptive approach improves the LaTeX output.

\section{Related work}
Recent DNN-based architectures have significantly improved the accuracy of HMER. These are mainly dominated by sequential encoder-decoder models (most of which are also based on Transformer architecture), including CoMER \cite{avidan_comer_2022}. Multi-task learning models further explore structural decoders and additional features to improve performance \cite{zhu_tamer_2025, leonardis_posformer_2025, zhang_ssan_2025}. One such feature is frequency domain information via the discrete cosine transform, incorporated in the MFH (Marries Frequency Domain with HMER) method \cite{lin_mfh_2025}. With this, MFH-CoMER achieves fair accuracy rates of 61.66\%-63.72\% on the CROHME datasets while maintaining structural simplicity.

Leading HMER models can be generally classified into those that focus on developing new architectures and those that fine-tune existing context-rich large language models---although both largely rely on attention-based mechanisms. Among the leading new architectures is Sunia, which translates the image of a math expression into a sequence of tokens where each math expression has a unique representation---using an encoder of stacked CNN/BLSTM layers and an attention-based decoder to model convergence and estimate each symbol location \cite{fink_icdar_2023}. This achieved best performance on the 2023 CROHME Competition on HMER. Other competitive HMER models include the YP\_OCR using attention-based encoder-decoder architecture with DenseNet encoder \cite{fink_icdar_2023} and TST using a Tree Structured Transformer \cite{yin_tst_2026}. For fine-tuning dependent HMER, also among the leading sequential encoder-decoder models is Uni-MuMER \cite{li_uni-mumer_2025}, developed by fine-tuning an existing VLM (Vision-Language Model), specifically the Qwen2.5-VL-3B \cite{bai_qwen25-vl_2025} that employs a redesigned Vision Transformer (ViT) architecture. It achieves state-of-the-art in multiple HMER benchmarks, reaching an average expression recognition rate of 79.74\% on the CROHME 2014, 2016, 2019 datasets \cite{li_uni-mumer_2025}. Other recent models use a similarly fine-tuning-based approach, e.g., La-Math-Ex fine-tuned from the TrOCR‑Base vision‑language Transformer \cite{coelho_-math-ex_2025}. However, these models are generally much larger than others, as they typically rely on models with parameter counts in the billions. Furthermore, due to sequential architectures generally being prone to cascading errors in its process, other approaches also continue to be explored. For instance, the discrete diffusion framework uses iterative symbolic refinement instead of left-to-right generation \cite{kawakatsu_gryphone_2026}. This outperforms previous state-of-the-art models, achieving a character error rate (CER) of 5.56\% and exact match accuracy of 60.42\% on the MathWriting benchmark, as well as increased performance on CROHME 2014-2023.

In many of these architectures, among the main issues that they try to address is visual confusion, mainly caused by ambiguously written expressions (which can cause inconsistent outputs \cite{kawakatsu_gryphone_2026}) and visually similar characters \cite{li_uni-mumer_2025}. Visually similar characters continue to be sometimes confused by recent architectures \cite{li_uni-mumer_2025, yin_return_2026}, especially models that rely mostly on structural methods \cite{yin_return_2026}. However, among the recent techniques aiming to reduce this confusion, there is no existing approach that considers the specific handwriting patterns of an individual writer. Applying writer-adapted fine-tuning for HMER remains largely unexplored, although similar handwriting tasks (e.g., handwritten text recognition \cite{huttner_low-rank_2025} and mathematics expression evaluation \cite{nguyen_vehme_2025}) have tested other fine-tuning applications with relative success. Still no current state-of-the-art-competitive systems adjust to the way a specific person writes. By building a writer adaptive system that learns an individual’s handwriting and outputs the written input into LaTeX directly, this study will then provide a practical tool that lowers the time and effort that it takes to digitize handwritten mathematical expressions while improving its accuracy for the people who use it. We aim to fill a gap in the field which is that writer adaptation has been shown to improve accuracy in handwritten text recognition, but no proponent has applied it to mathematical expressions.
%

\section{Methodology}
\subsection{Preliminaries}

\subsubsection{Task Formulation and Notation} \label{sec:notation}

Given an input binarized handwritten image $\mathbf{X} \in \{0, 1\}^{H \times W}$, we define the HMER task as predicting a target \LaTeX{} token sequence $\mathbf{Y} = (y_1, y_2, \dots, y_L)$ of length $L$ where any $y_i \in \mathbf{Y}$ is a member of a fixed vocabulary $\mathcal{V}$, i.e., $y_i \in \mathcal{V}$. More formally:
\begin{equation}
    P(\mathbf{Y} | \mathbf{X}) = \prod_{t=1}^T P(y_t | y_{<t}, \mathbf{X})
\end{equation}
which is then optimized as the following negative log-likelihood:
\begin{equation}
    \mathcal{L} = - \sum_{t = 1}^T \log P(y_t | y_{<t}, \mathbf{X})
\end{equation}



\subsubsection{Discrete Cosine Transform} \label{sec:dct}

MFH-CoMER \cite{lin_mfh_2025} introduces frequency-domain information to the CoMER architecture. Specifically, images are divided into patches $f(x, y) \in \mathbb{R}^{n \times n}$ of size $n \times n$. Each patch is then transformed into a table of frequency component coefficients $F(u, v) \in \mathbb{R}^{n \times n}$ using the discrete cosine transform defined as:
\begin{equation}
\begin{split}
  F(u, v) = \frac{2}{n} C(u) C(v) \sum_{x=0}^{n-1} \sum_{y=0}^{n-1} & f(x, y) \cos \left[ \frac{(2x + 1)u\pi}{2n} \right] \\
  & \times \cos \left[ \frac{(2y + 1)v\pi}{2n} \right]
\end{split}
\end{equation}
%
where the scaling factor $C(\alpha)$, for normalization, is given by:
\begin{equation}
    C(\alpha) = \begin{cases}
        \displaystyle \frac{1}{\sqrt{2}}, & \text{if } \alpha = 0 \\ 
        1, & \text{otherwise} 
    \end{cases}
\end{equation}
Note this means that $(u, v)$ and $(x, y)$ are in some \textit{numeric} correspondence with each other as $F$ serves as the frequency-domain representation of the spatial-domain information in $f$ (though note that $(u, v)$ does \textit{not} map to a spatial location $(x, y)$). $F(u, v)$ is instead a \textit{scalar} weight that represents the frequency pattern characterized by $u$ horizontal and $v$ vertical oscillations across the entire patch. Thus $F(0, 0)$ is the DC component featuring no oscillations, while the highest-frequency AC component is given by $F(n - 1, n - 1)$.

\subsection{Image Pre-Processing}

\subsubsection{Otsu's Method for Binarization} 

Let $\mathbf{I} \in \{0, 1, \dots, 255\}^{H \times W}$ be an input raw grayscale image. Binarization refers to transforming this image into the black-and-white input image $\mathbf{X} \in \{0, 1\}^{H \times W}$ described in section~\ref{sec:notation}. Equivalently, we must define some function $f : \{0, 1, \dots, 256\} \to \{0, 1\}$ and apply it to every pixel intensity value $\mathbf{I}(x, y)$ in the image. Otsu's method \cite{otsu_threshold_1979} non-parametrically determines some optimal intensity threshold $t^*$ to determine the output value for a pixel:
\begin{equation}
    f(\mathbf{I}(x, y)) = \begin{cases}
        1, & \text{if } \mathbf{I}(x, y) > t^* \\
        0, & \text{otherwise}
    \end{cases}
\end{equation}
In other words, each pixel $\mathbf{I}(x, y)$ is assigned to one of two classes: the \textit{background} pixels $C_0$ for which $f(\mathbf{I}(x, y)) = 0$ (i.e., $\mathbf{I}(x, y) \in [0, t]$) and the \textit{foreground} pixels $C_1$ for which $f(\mathbf{I}(x, y)) = 1$ (i.e., $\mathbf{I}(x, y) \in [t+1, 255]$). Otsu's method must then select $t^*$ such that it maximizes the inter-class variance:
\begin{equation}
    t^* = \arg \max \omega_0(t) \omega_1(t) (\mu_0(t) - \mu_1(t))^2
\end{equation}
where $\omega_0(t)$ and $\omega_1(t)$ refer to the cumulative probabilities for each class while $\mu_0(t)$ and $\mu_1(t)$ refer to their mean intensities.


\subsection{Base MFH-CoMER}

\begin{figure}
    \centering
    \includegraphics[width=\linewidth]{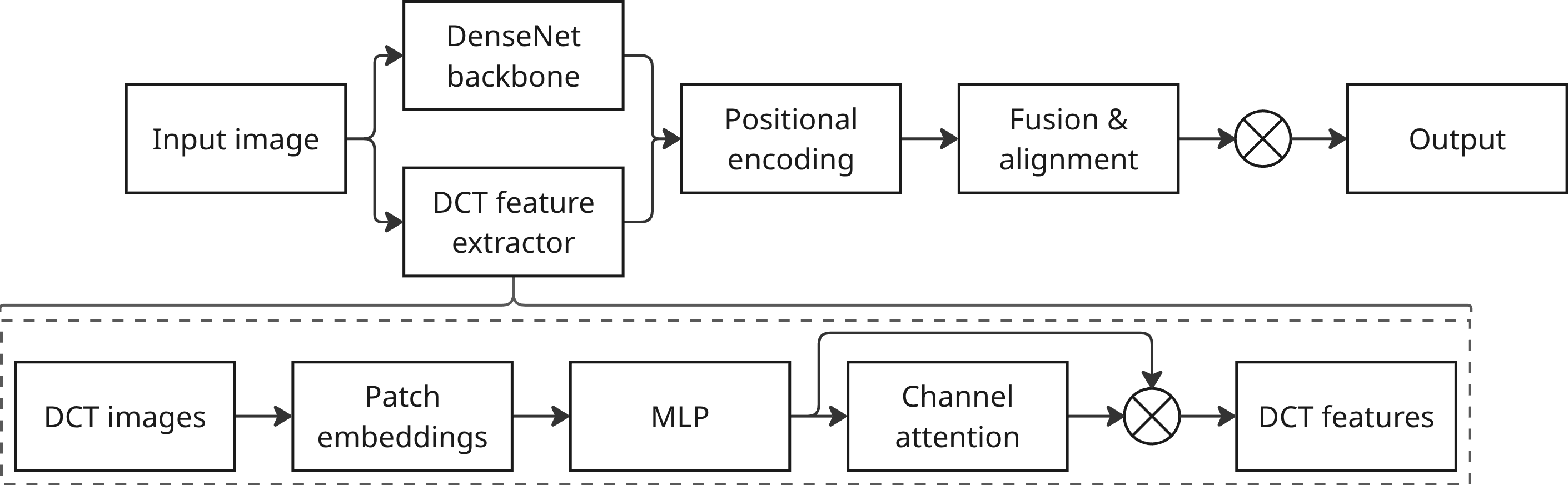}
    \caption{MFH-CoMER architecture \cite{lin_mfh_2025}}
    \label{fig:mfh-comer}
\end{figure}

MFH-CoMER \cite{lin_mfh_2025} delegates the processing of frequency-domain information to a separate feature extractor module (see \autoref{fig:mfh-comer}). Images are processed with the DCT with a patch size of $n=8$ (see section~\ref{sec:dct}) to generate a set of frequency coefficient tables $F(u, v)$, with each table thus representing an $8 \times 8$ patch. For each patch, \cite{lin_mfh_2025} notes that high-frequency information helps capture the contours of the handwritten expressions, meaning the feature extraction module must preserve these frequencies while suppressing low-frequency components. As such, all frequency coefficients outside the bottom-right $m \times m$ area (where $m \leq n)$ are set to zero. Transformed patches are then transformed back into the spatial domain to produce the DCT images processed by the rest of the module. Each patch is tokenized, processed by MLP layers and channel-wise attention mechanisms to produce the required DCT features fed into the rest of the model.

\subsection{Author-Specific Adaptation}

Full-parameter fine-tuning with author-specific data is applied to the base pretrained MFH-CoMER model to obtain each author-specific adaptation of the model. All parameters in the encoder, attention, decoder, and embeddings are unfrozen and fine-tuned using a fresh AdamW optimizer with lowered learning rate (1e-5) and early stopping (patience of 5) on validation expression recognition rate, up to a maximum of 20 epochs. The checkpoint with best validation performance for each author-specific adaptation is used for evaluation. This low-resource light fine-tuning is in consideration of the relatively smaller training sets used.

Expression recognition rate (ExpRate) and character error rate (CER) are used as evaluation metrics. ExpRate is defined as the percentage of exact matches to the expected LaTeX expression. CER is computed as the Levenshtein distance to turn the predicted equation into the expected equation. Given $S$ substitutions, $D$ deletions, and $I$ insertions needed, and $N$ total characters in the expected equation, CER is more formally defined as:
\begin{equation}
    CER = \frac{S + D + I}{N}
\end{equation}
Given the results from multiple runs via five training seeds for data shuffling per author-specific adaptation (i.e., the fine-tuning images are fixed, but the order in which they are shuffled is altered per seed) evaluated against a fixed author-specific test set, following the setup outlined in section~\ref{sec:datasets} to account for limited data, the following are performed for further analysis:

\subsubsection{Wilson score interval}

A confidence interval for the binomial proportion $\hat{p}=k/n$ with $k$ exact matches out of $n$ test items is computed for the base model and for every adapted model from each individual seed. This is to quantify the ``uncertainty" of the test-set sampling for the reported ExpRates, in consideration of the relatively small fine-tuning sets used. This can be formally expressed as:
\begin{equation}
    CI_{95\%} = \frac{\hat{p} + \frac{z^2}{2n} \pm z \sqrt{\frac{\hat{p}(1 - \hat{p})}{n} + \frac{z^2}{4n^2}}}{1 + \frac{z^2}{n}}, \quad z = 1.96
\end{equation}

\subsubsection{Exact McNemar's test}
\label{sec:mcnemar}

This is used in consideration of the setup given limited data, with precedent for comparisons in image processing tasks with ``pass"/``fail" and ``correct"/``incorrect" outcomes \cite{li_homography_2023, militello_fingerprint_2021}, 
analogous to the exact match or inexact match result that determines ExpRate, our primary metric. For each training seed; i.e., for each of five independently trained models for a given writer adaptation with the same author set and train size, the paired test compares the baseline and adapted model. This is used over standard chi-square approximation due to the smaller per-author test sets. Given pairs $n_{01}$ (for items that the base model predicted incorrectly but the adapted model predicted correctly) and $n_{10}$ (for items that the base model predicted correctly but the adapted model predicted incorrectly), under the null hypothesis that the two models have identical accuracy, expressed as $H_0 : n_{01}, n_{10} \sim \text{Binomial}(n_{01} + n_{10}, 0.5)$, the two-sided exact $p$-value is more formally defined as:
%
\begin{equation}
\begin{aligned}
  p &= 2 \min \left( 1, \sum_{i=0}^{k} \binom{n}{i} 0.5^n \right), \\[4pt]
  \text{where } n &= n_{01} + n_{10}, \quad k = \min(n_{01}, n_{10})
\end{aligned}
\end{equation}

\subsubsection{Item-level permutation test}
\label{sec:item-level-permutation}

Since the five training seeds are evaluated against the same fixed test set, per-item correctness is correlated across seeds; an item that is ``difficult", or that the baseline model already struggles with, tends to produce a similar discordance pattern regardless of which seed is used for the adapted model. This is a case of the \textit{dependent} observation issue that is more commonly found in NLP settings, as standard significance tests generally assume \textit{independent} samples \cite{dror-etal-2018-hitchhikers}. 

Since no existing significance test in the surveyed literature is specifically designed for this pooled item-by-seed dependency structure, an item-level permutation test is formulated to directly use pooled item-by-seed data, following sign-randomization construction described in  \cite[\S\S2.2.2, 3.2.1]{good_permutation_2005}. For test item $j \in \{1,\dots,N\}$ $(N=46)$ and seed $s \in \{1,\dots,S\}$ $(S=5)$, let $b_j \in \{0,1\}$ represent whether the prediction of the baseline for item $j$ is correct, and $a_{j,s} \in \{0,1\}$ represent whether the prediction of the $s$-seed adapted model for item $j$ is correct. The per-cell discordance is defined as:

\begin{equation}
    \delta_{j,s} = a_{j,s} - b_j \in \{-1, 0, +1\}
\end{equation}

From this, $+1$ would represent an item that adapted got correct but baseline got wrong, $-1$ would represent the reverse, and $0$ would mean both models agreed. Pooling across all items and seeds, the test statistic is defined as:

\begin{equation}
    T_{\text{obs}} = \sum_{j=1}^{N} \sum_{s=1}^{S} \delta_{j,s} = n_{01} - n_{10}
\end{equation}
using $n_{01}$, $n_{10}$ as defined in \autoref{sec:mcnemar} for the same directional quantity McNemar's test targets, now pooled across seeds.

Under $H_0$ (adaptation truly has no effect on correctness), for any item where the base model and adapted model disagree, there is no real basis as to why adapted would be correct or incorrect, so this would be analogous to a ``fair coin flip", and roughly equal numbers of $n_{01}$ and $n_{10}$ can be expected. Since $b_j$ is the same value across all $S$ seeds for item $j$, the five outcomes for that item are not independent of one another; treating each $(j,s)$ cell independently would understate true variance and inflate significance. Instead, to avoid this, each item's full row of outcomes is treated as a single unit of exchangeability. For each item $j$, the item-level row can be defined as:

\begin{equation}
    \mathbf{d}_j = (\delta_{j,1}, \delta_{j,2}, \dots, \delta_{j,S})
\end{equation}

From this, a cluster (item-block) sign-flip permutation can be performed by jointly sign-flipping the entire row rather than flipping each individual $(j, s)$ cell independently.

For each of $M$ Monte Carlo draws, we sample independent Rademacher signs $\varepsilon_j^{(m)} \in \{-1, +1\}$ per item and compute the statistic:

\begin{equation}
    T^{(m)} = \sum_{j=1}^{N} \varepsilon_j^{(m)} \sum_{s=1}^{S} \delta_{j,s}, \quad m = 1, \dots, M
\end{equation}

Using $M = 100000$ draws, the two-sided permutation p-value is calculated as the fraction of permuted statistics at least as extreme as the one actually observed. This is defined as:

\begin{equation}
    p = \frac{1}{M} \sum_{m=1}^{M} 1 \left[ \left| T^{(m)} \right| \ge \left| T_{\text{obs}} \right| \right]
\end{equation}

\subsection{Domain Gap Analyses}

The original MFH-CoMER model was trained on digital, binary images. This means that it has not seen real-world images at any point during training. There thus exists a domain gap between the model's training data and the images that it will see in practice during inference time; for instance, since the model was trained on perfectly binary images with static pen weights, the natural variations in stroke width---as well as artifacts like blur---present in physical images might slightly damage performance even after the binarization process. We employ the following analyses to characterize this domain gap.



\subsubsection{Fragmentation}

We borrow the formalization from \cite{rosenfeld_sequential_1966}. Under an 8-connectivity metric, two distinct foreground pixels $p_1 = (x_1, y_1), p_2 = (x_2, y_2) \in C_1$ are defined as \textit{adjacent} if:
\begin{equation}
    \max(|x_1 - x_2|, |y_1 - y_2|) = 1
\end{equation}
Two pixels $p_1, p_2 \in C_1$ can be said to have a \textit{path} between them if there exists some sequence $(q_0, q_1, \dots, q_k) \subseteq C_1$ such that $q_0 = p_1$, $q_k = p_2$, and, for all $1 \leq i \leq k$, $q_i$ is adjacent to $q_{i - 1}$. In this case we say that $p_1$ and $p_2$ are \textit{connected} (note that obviously any two pixels that are adjacent are also connected). This partitions $C_1$ into $N_{\text{cc}}$ mutually disjoint subsets $\{K_1, K_2, \dots, K_{N_{\text{cc}}}\}$, referred to as \textit{connected components}, where all pixels within $K_i$ are connected to one another. We report the raw count of connected components as a measure of the fragmentation that results from pen ink discontinuities or image and artifacts.

Furthermore, to better distinguish image noise from true stroke differences, we report also the mean pixel area of each component $A(K_i) = |K_i|$ as well as the component density $D_{\text{ink}}$, defined as the number of components divided by the number of foreground pixels, i.e.:
\begin{equation}
    D_{\text{ink}} = \frac{N_{\text{cc}}}{|C_1|}
\end{equation}
We also report the fraction of \textit{noise specks} $f_{\text{speck}}$, computed as the fraction of components that fall below a selected area threshold $\tau_\text{speck}$:
\begin{equation}
f_{\text{speck}} = \frac{|\{K_i \mid A(K_i) \le \tau_{\text{speck}}\}|}{N_{\text{cc}}}
\end{equation}
where for all experiments, $\tau_{\text{speck}} = 2 \text{ px}$.

\subsubsection{Stroke width}

To quantify possible changes in stroke width between digital and physical media, we apply the \textit{Eucledian Distance Transform} (EDT) to all pixels in the foreground $C_1$. For any pixel $p \in C_1$, the EDT calculates the minimum distance to the nearest pixel in the background $C_0$. We evaluate stroke width strictly on interior stroke pixels on a 1-pixel skeleton extracted via \texttt{scikit-image} (or \texttt{bwmorph} in MATLAB). We summarize stroke width by the sample mean and coefficient of variation $\text{CV}_W$ of the resulting EDT values. Higher $\text{CV}_W$ values capture ink flow irregularities and binarization edge jitter along stroke trajectories.

\section{Experimental setup}

\subsection{Datasets}
\label{sec:datasets}

\subsubsection{CROHME}

The Competition on Recognition of Online Handwritten Mathematical Expressions (CROHME) \cite{fink_icdar_2023} offers a dataset for HMER that is often used as a baseline for model training. Four dataset splits are publicly available: the training set plus each of the evaluation sets for the 2014, 2016, and 2019 iterations of the competition. For author-specific adaptation evaluation, we use the 2014 and 2016 split divided per author identifier for all handwriting samples by an author who contributed at least 20 handwriting samples. As previously mentioned, all images are in a fully binarized bitmap format with a uniform stroke thickness. We apply a 50\%/30\%/20\% test/fine-tune/validation split due to limited data and run each author adaptation with 5 different seeds for data shuffling, and we report the resulting mean and standard variation. The token inventory for this dataset is left unmodified for all other datasets (equations violating it are removed).

\subsubsection{Custom digital handwriting}

\label{sec:dig-handwriting}

\begin{figure}[h]
	\centering
	\begin{subfigure}[h]{0.7\linewidth}
		\centering
		\includegraphics[width=\linewidth]{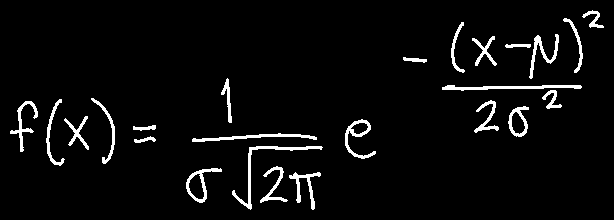}
		\caption{Digitally-handwritten image with a static pen weight. Imitates the CROHME dataset.}
		\label{fig:digital}
	\end{subfigure}
    
	\begin{subfigure}[h]{0.7\linewidth}
		\centering
		\includegraphics[width=\linewidth]{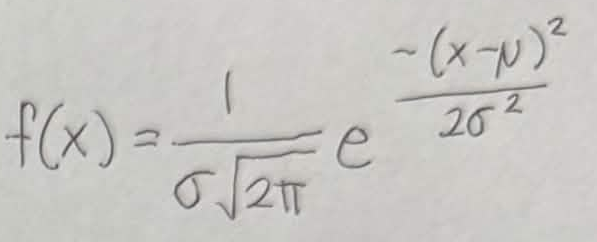}
		\caption{Image of handwritten equation on paper, with no processing applied.}
		\label{fig:physical-raw}
	\end{subfigure}   
    
	\begin{subfigure}[h]{0.7\linewidth}
		\centering
		\includegraphics[width=\linewidth]{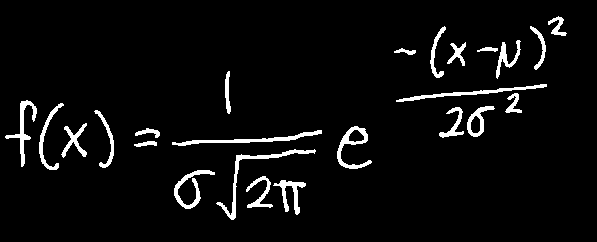}
		\caption{Binarized version of the image in \autoref{fig:physical-raw} using Otsu's method.}
		\label{fig:physical-binarized}
	\end{subfigure}
	\caption{Sample images from the custom-author dataset. These images (both digital and physical) are all from a single author and correspond to the mathematical expression $\displaystyle f(x) = \frac{1}{\sigma \sqrt{2\pi}} e^{-\frac{(x - \mu)^2}{2\sigma^2}}$.}
    
	\label{fig:handwriting-samples}
\end{figure}

To further evaluate the author-specific adaptation mechanism, we also introduce two new evaluation sets each---one digital and one physical--for two distinct authors. Each set comprises 71 arbitrarily-selected equations of varying complexity. The digital set was generated using makeshift digital application. The final bitmap images have stroke width uniform across an author subset, similar to the sample images in CROHME. An example of one of the digital image samples may be seen in \autoref{fig:digital}. Of the 71 sample images, we use 46 as a held-out test set and five for validation. We then perform ablations with the number of images $n$ used for fine-tuning, where $n \in \{5, 8, 11, 14, 17, 20 \}$ to find the extent to which fine-tuning on more data can improve or worsen the model's performance relative to the baseline. Each ablation is tested with 5 different seeds for data shuffling (with dataset splits kept strictly identical), and we report the resulting mean and standard deviation.

\subsubsection{Custom physical handwriting} 
\label{sec:phys-handwriting}

The physical set, on the other hand, was generated by writing the equations on a clean sheet of paper, taking a photo using a smartphone camera, and then segmenting the images. The raw segmented images (see \autoref{fig:physical-raw}) are then binarized using Otsu's method to produce images that are more similar to the digital set (and by extension CROHME; see \autoref{fig:physical-binarized}). The physical handwriting dataset is further classified into standard handwriting for two distinct authors and rushed handwriting for one author. Rushed handwriting is simulated via writing the same 71 images within 7 minutes, with an additional 30 seconds for checking and correcting any errors. We split the 71 images identically and perform the same ablation tests as with the digital images.




\section{Results and discussion}
\subsection{Image Analyses}

\begin{table}[htbp]
\centering
\caption{Comparison of stroke structure metrics between digital and physical images.
$N_{\text{cc}}$: component count;
$F_{\text{speck}}$: speck fraction ($\le \tau_{\text{speck}}$);
$\bar{A}$: mean area (px);
$D_{\text{ink}}$: ink density (/1k px);
$\bar{W}$: stroke width mean (px);
$\text{CV}_W$: stroke width coefficient of variation.
Dig.\ = Digital, Phys.\ = Physical (mean $\pm$ std); $r$ = Pearson correlation; $p$ = Wilcoxon $p$-value.}
\label{tab:stroke_metrics_raw_combined}

\begin{subtable}{\linewidth}
\centering
\caption{Author 0: digital vs.\ no-rush physical images}
\label{subtab:author0_norush}
\footnotesize
\setlength{\tabcolsep}{3pt}
\begin{tabularx}{\linewidth}{l|X X r r}
\toprule
Metric & Dig. (mean $\pm$ std) & Phys. (mean $\pm$ std) & $r$ & $p$ \\
\midrule
$N_{\text{cc}}$ & $10.27 \pm 6.78$ & $\uparrow 10.49 \pm 6.75$ & $0.992$ & $0.0312^{*}$ \\
$F_{\text{speck}}$ & $0.000 \pm 0.000$ & $\uparrow 0.012 \pm 0.041$ & - & $0.0117^{*}$ \\
$\bar{A}$ & $355.37 \pm 112.57$ & $\downarrow 276.53 \pm 96.11$ & $0.593$ & $<0.001^{*}$ \\
$D_{\text{ink}}$ & $3.08 \pm 0.92$ & $\uparrow 3.94 \pm 1.03$ & $0.531$ & $<0.001^{*}$ \\
$\bar{W}$ & $2.68 \pm 0.03$ & $\uparrow 2.71 \pm 0.15$ & $0.284$ & $0.0402^{*}$ \\
$\text{CV}_W$ & $0.334 \pm 0.012$ & $\uparrow 0.342 \pm 0.022$ & $0.400$ & $<0.001^{*}$ \\
\bottomrule
\end{tabularx}
\end{subtable}

\medskip

\begin{subtable}{\linewidth}
\centering
\caption{Author 1: digital vs.\ no-rush physical images}
\label{subtab:author1_norush}
\footnotesize
\setlength{\tabcolsep}{3pt}
\begin{tabularx}{\linewidth}{l|X X r r}
\toprule
Metric & Dig. (mean $\pm$ std) & Phys. (mean $\pm$ std) & $r$ & $p$ \\
\midrule
$N_{\text{cc}}$ & $9.34 \pm 5.98$ & $\uparrow 10.07 \pm 6.71$ & $0.986$ & $<0.001^{*}$ \\
$F_{\text{speck}}$ & $0.001 \pm 0.009$ & $\uparrow 0.004 \pm 0.016$ & $0.254$ & $0.3452^{}$ \\
$\bar{A}$ & $121.37 \pm 49.12$ & $\downarrow 63.16 \pm 19.89$ & $0.416$ & $<0.001^{*}$ \\
$D_{\text{ink}}$ & $9.57 \pm 3.70$ & $\uparrow 17.25 \pm 4.90$ & $0.225$ & $<0.001^{*}$ \\
$\bar{W}$ & $2.05 \pm 0.02$ & $\downarrow 2.04 \pm 0.03$ & $0.038$ & $0.2054^{}$ \\
$\text{CV}_W$ & $0.112 \pm 0.034$ & $\downarrow 0.101 \pm 0.041$ & $0.127$ & $0.0299^{*}$ \\
\bottomrule
\end{tabularx}
\end{subtable}

\medskip

\begin{subtable}{\linewidth}
\centering
\caption{Author 1: digital vs.\ rushed physical images}
\label{subtab:author1_rush}
\footnotesize
\setlength{\tabcolsep}{3pt}
\begin{tabularx}{\linewidth}{l|X X r r}
\toprule
Metric & Dig. (mean $\pm$ std) & Phys. (mean $\pm$ std) & $r$ & $p$ \\
\midrule
$N_{\text{cc}}$ & $9.34 \pm 5.98$ & $\uparrow 9.90 \pm 6.39$ & $0.984$ & $<0.001^{*}$ \\
$F_{\text{speck}}$ & $0.001 \pm 0.009$ & $\uparrow 0.009 \pm 0.027$ & $0.430$ & $0.0109^{*}$ \\
$\bar{A}$ & $121.37 \pm 49.12$ & $\downarrow 58.36 \pm 17.17$ & $0.363$ & $<0.001^{*}$ \\
$D_{\text{ink}}$ & $9.57 \pm 3.70$ & $\uparrow 18.44 \pm 4.81$ & $0.309$ & $<0.001^{*}$ \\
$\bar{W}$ & $2.05 \pm 0.02$ & $\downarrow 2.04 \pm 0.03$ & $0.022$ & $0.0104^{*}$ \\
$\text{CV}_W$ & $0.112 \pm 0.034$ & $\downarrow 0.092 \pm 0.039$ & $0.002$ & $<0.001^{*}$ \\
\bottomrule
\end{tabularx}
\end{subtable}

\end{table}

Findings for all three custom-author dataset are summarized in \autoref{tab:stroke_metrics_raw_combined}. The behaviors of the first four metrics (namely component count, speck fraction, mean component area, and ink density) were consistent across all sets. All metrics, except mean component area, increased for the physical sets compared to the digital sets. Notably, mean component area was almost halved for both of Author 1's physical sets.


\subsection{CROHME (2014, 2016)}

Across all authors with at least 20 handwriting samples (with none found in 2019), the average ExpRate decreased by 0.37\% compared to base model ExpRate. This is largely due to the limited dataset, consisting of only 7 fine-tuning samples per author for 61 of the 64 evaluated authors. The statistical power of the test set for each adaptation is similarly limited by lack of data. This suggests that more author data is needed for the task of writer-adaptive HMER, both with more fine-tuning samples for possibly improved fine-tuning and more test samples to be able to establish statistical significance. These are largely explored through the custom handwriting datasets that follow.

\subsection{Custom digital handwriting}

For both authors, \autoref{table:digi-handwriting} shows that fine-tuned writer-adapted models have a relatively consistent increase in both ExpRate and accuracies for predictions with ``$\leq1$ Error" and ``$\leq2$ Error", with all recorded mean values greater than base model performance. The writer-adapted models also show a decrease in average CER that remains consistently lower than base model CER.

\begin{table*}[h]
\centering
\caption{ Results on digital handwriting datasets. For ``$\leq$1 Error", ``$\leq$2 Error", an error is defined as a LaTeX symbol insertion, deletion, or substitution. All results outperform or are equal to base model performance metric. }
\label{table:digi-handwriting}
\begin{tabularx}{\textwidth}{l|r|XX|XX|XX|XX}
\toprule
\multirow{2}{*}{\textbf{Handwriting}} & \multicolumn{1}{c|}{\multirow{2}{*}{\textbf{Fine-tune size}}} & \multicolumn{2}{c|}{\textbf{ExpRate}} & \multicolumn{2}{c|}{\textbf{$\leq$1 Error}} & \multicolumn{2}{c|}{\textbf{$\leq$2 Error}} & \multicolumn{2}{c}{\textbf{CER}} \\ \cline{3-10} 
 & \multicolumn{1}{c|}{} & \multicolumn{1}{c}{\textbf{mean}} & \multicolumn{1}{c|}{\textbf{std}} & \multicolumn{1}{c}{\textbf{mean}} & \multicolumn{1}{c|}{\textbf{std}} & \multicolumn{1}{c}{\textbf{mean}} & \multicolumn{1}{c|}{\textbf{std}} & \multicolumn{1}{c}{\textbf{mean}} & \multicolumn{1}{c}{\textbf{std}} \\ \midrule
\multirow{7}{*}{Author 0} & base & 76.09\% & - & 93.48\% & - & 97.83\% & - & 2.52\% & - \\ \cline{2-10} 
 & 5 & 78.26\% & 0.00\% & 96.52\% & 1.19\% & 97.83\% & 0.00\% & 1.84\% & 0.05\% \\
 & 8 & 81.30\% & 1.19\% & 97.83\% & 2.17\% & 98.70\% & 1.19\% & 1.52\% & 0.31\% \\
 & 11 & 83.04\% & 2.83\% & 98.70\% & 1.94\% & 99.13\% & 1.19\% & 1.34\% & 0.34\% \\
 & 14 & 83.48\% & 1.94\% & 99.13\% & 1.19\% & 99.13\% & 1.19\% & 1.30\% & 0.26\% \\
 & 17 & 83.91\% & 1.94\% & 97.83\% & 1.54\% & 97.83\% & 1.54\% & 1.47\% & 0.29\% \\
 & 20 & 84.78\% & 1.54\% & 99.13\% & 1.19\% & 99.13\% & 1.19\% & 1.23\% & 0.26\% \\ \midrule
\multirow{7}{*}{Author 1} & base & 50.00\% & - & 69.57\% & - & 84.78\% & - & 5.56\% & - \\ \cline{2-10} 
 & 5 & 63.04\% & 0.00\% & 76.09\% & 0.00\% & 91.30\% & 0.00\% & 3.48\% & 0.00\% \\
 & 8 & 52.17\% & 0.00\% & 69.57\% & 0.00\% & 86.52\% & 1.82\% & 4.94\% & 0.14\% \\
 & 11 & 60.00\% & 1.19\% & 73.91\% & 0.00\% & 91.30\% & 0.00\% & 3.62\% & 0.05\% \\
 & 14 & 62.17\% & 1.19\% & 77.39\% & 1.94\% & 91.30\% & 0.00\% & 3.57\% & 0.26\% \\
 & 17 & 62.17\% & 2.48\% & 81.30\% & 1.19\% & 92.17\% & 1.94\% & 3.72\% & 0.34\% \\
 & 20 & 63.48\% & 2.83\% & 81.30\% & 1.94\% & 92.17\% & 1.94\% & 3.74\% & 0.42\% \\ 
 \bottomrule
\end{tabularx}
\end{table*}

The Wilson 95\% confidence intervals (at test set n=46, which is unchanged across seeds and sizes for all runs) are considerably wide with ExpRate Wilson CI of 0.6206-0.8609 for Author 0 baseline model and 0.3612-0.6388 for Author 1 baseline model, with the respective adapted models having similarly wide intervals. As such, significance is assessed via the McNemar test for the ablations with fine-tuning size of 20. In \autoref{tab:mcnemar_results}, for both authors across five seeds, the results show consistent direction in that the adapted model has better performance than base model. Although most individual $p$-values from the McNemar tests do not reach significance,
the item-level permutation test results suggest consistent directionality for Author 0 at $p < 0.05$ (reflecting how no previously correct items from Author 0 turned incorrect after model adaptation), but less consistency for Author 1 due to item-level noise.

\begin{table}[h]
\centering
\caption{ Results on McNemar's test performed on digital handwriting-adapted models across 5 seeds. Results on item-level permutation test are below each author table. }
\label{tab:mcnemar_results}

\begin{subtable}[t]{\linewidth}
\centering
\begin{tabularx}{\linewidth}{c|c c c X}
\toprule
\textbf{Seed} & $n_{01}$ & $n_{10}$ & $p$-value & \textbf{Direction} \\
\midrule
(1) & 3 & 0 & 0.2500 & adapted better \\
\hline
(2) & 5 & 0 & 0.0625 & adapted better \\
\hline
(3) & 4 & 0 & 0.1250 & adapted better \\
\hline
(4) & 4 & 0 & 0.1250 & adapted better \\
\hline
(5) & 4 & 0 & 0.1250 & adapted better \\
\bottomrule
Permutation test & 20 & 0 & 0.0316 & \\
\end{tabularx}
\caption{Author 0}
\end{subtable}

\begin{subtable}[t]{\linewidth}
\centering
\begin{tabularx}{\linewidth}{c|c c c X}
\toprule
\textbf{Seed} & $n_{01}$ & $n_{10}$ & $p$-value & \textbf{Direction} \\
\midrule
(1) & 10 & 2 & 0.0386 & adapted better \\
\hline
(2) & 9 & 4 & 0.2668 & adapted better \\
\hline
(3) & 9 & 2 & 0.0654 & adapted better \\
\hline
(4) & 9 & 4 & 0.2668 & adapted better \\
\hline
(5) & 8 & 2 & 0.1094 & adapted better \\
\bottomrule
Permutation test & 45 & 14 & 0.0670 & \\
\end{tabularx}
\caption{Author 1}
\end{subtable}

\end{table}

\subsection{Custom physical handwriting}

\autoref{table:phys-handwriting} shows less consistent results for fine-tuning on physical handwriting. The models adapted to the standard physical handwriting of Author 0 shows a slight decrease in average ExpRate and slight increase in average CER compared to base model performance. Meanwhile, models fine-tuned on Author 1 physical handwriting show an increase in average ExpRate and general decrease in average CER. 

\begin{table*}[h]
\centering
\caption{ Results on physical handwriting datasets. $\downarrow$ indicates lower ExpRate, ``$\leq$1 Error", or ``$\leq$2 Error" compared to base model metric. $\uparrow$ indicates higher CER compared to base model CER. All other results outperform or are equal to base model performance metric. }
\label{table:phys-handwriting}
\begin{tabularx}{\textwidth}{l|r|XX|XX|XX|XX}
\toprule
\multirow{2}{*}{\textbf{Handwriting}} & \multicolumn{1}{c|}{\multirow{2}{*}{\textbf{Fine-tune size}}} & \multicolumn{2}{c|}{\textbf{ExpRate}} & \multicolumn{2}{c|}{\textbf{$\leq$1 Error}} & \multicolumn{2}{c|}{\textbf{$\leq$2 Error}} & \multicolumn{2}{c}{\textbf{CER}} \\ \cline{3-10} 
 & \multicolumn{1}{c|}{} & \multicolumn{1}{c}{\textbf{mean}} & \multicolumn{1}{c|}{\textbf{std}} & \multicolumn{1}{c}{\textbf{mean}} & \multicolumn{1}{c|}{\textbf{std}} & \multicolumn{1}{c}{\textbf{mean}} & \multicolumn{1}{c|}{\textbf{std}} & \multicolumn{1}{c}{\textbf{mean}} & \multicolumn{1}{c}{\textbf{std}} \\ \midrule
\multirow{7}{*}{Author 0 - standard} & base & 91.30\% & - & 95.65\% & - & 100.00\% & - & 1.23\% & - \\ \cline{2-10} 
 & 5 & $\downarrow$ 88.70\% & 0.97\% & $\downarrow$ 95.22\% & 1.82\% & 100.00\% & 0.00\% & $\uparrow$ 1.58\% & 0.22\% \\
 & 8 & $\downarrow$ 89.57\% & 0.97\% & 95.65\% & 1.54\% & 100.00\% & 0.00\% & $\uparrow$ 1.51\% & 0.23\% \\
 & 11 & 91.30\% & 0.00\% & 97.83\% & 0.00\% & 100.00\% & 0.00\% & 1.11\% & 0.00\% \\
 & 14 & $\downarrow$ 90.87\% & 0.97\% & 96.96\% & 1.19\% & 100.00\% & 0.00\% & 1.23\% & 0.21\% \\
 & 17 & $\downarrow$ 90.87\% & 0.97\% & 97.83\% & 0.00\% & 100.00\% & 0.00\% & 1.13\% & 0.04\% \\
 & 20 & $\downarrow$ 89.57\% & 0.97\% & 96.52\% & 1.94\% & 100.00\% & 0.00\% & $\uparrow$ 1.31\% & 0.24\% \\ \midrule
\multirow{7}{*}{Author 1 - standard} & base & 67.39\% & - & 91.30\% & - & 93.48\% & - & 4.43\% & - \\ \cline{2-10} 
 & 5 & 71.74\% & 0.00\% & $\downarrow$ 89.13\% & 0.00\% & 95.65\% & 0.00\% & $\uparrow$ 4.44\% & 0.00\% \\
 & 8 & 72.61\% & 1.19\% & 93.04\% & 0.97\% & 95.22\% & 0.97\% & $\uparrow$ 4.45\% & 0.37\% \\
 & 11 & 76.09\% & 1.54\% & 92.61\% & 1.94\% & 96.96\% & 1.94\% & 3.16\% & 1.52\% \\
 & 14 & 75.65\% & 1.82\% & 91.30\% & 1.54\% & 95.65\% & 1.54\% & 2.24\% & 0.44\% \\
 & 17 & 80.00\% & 2.38\% & 95.65\% & 1.54\% & 99.57\% & 0.97\% & 1.05\% & 0.09\% \\
 & 20 & 77.83\% & 1.82\% & 95.65\% & 0.00\% & 100.00\% & 0.00\% & 1.09\% & 0.03\% \\ \midrule
\multirow{7}{*}{Author 1 - rushed} & base & 58.70\% & - & 86.96\% & - & 93.48\% & - & 4.07\% & - \\ \cline{2-10} 
 & 5 & 63.04\% & 0.00\% & $\downarrow$ 86.52\% & 0.97\% & 93.48\% & 0.00\% & $\uparrow$ 4.21\% & 0.08\% \\
 & 8 & 60.87\% & 0.00\% & 86.96\% & 0.00\% & 93.48\% & 0.00\% & $\uparrow$ 4.11\% & 0.07\% \\
 & 11 & 65.22\% & 0.00\% & 89.13\% & 0.00\% & 95.65\% & 0.00\% & 3.36\% & 0.03\% \\
 & 14 & 68.26\% & 1.94\% & 87.83\% & 1.19\% & 94.78\% & 1.19\% & 3.01\% & 0.18\% \\
 & 17 & 68.70\% & 1.19\% & 87.39\% & 1.82\% & 94.78\% & 1.19\% & 2.83\% & 0.10\% \\
 & 20 & 66.52\% & 1.19\% & $\downarrow$ 84.78\% & 1.54\% & $\downarrow$ 91.30\% & 0.00\% & 3.44\% & 0.16\% \\
 \bottomrule
\end{tabularx}
\end{table*}

Similar to findings in custom digital handwriting, the ExpRate Wilson 95\% confidence intervals are considerably wide for both the baselines (with 0.7968-0.9657 for Author 0 standard, 0.5297-0.7913 for Author 1 standard, and  0.4434-0.7171 for Author 1 rushed) and fine-tuned models. As such, significance is again assessed via the McNemar test for ablations with fine-tuning size of 20, as shown in \autoref{tab:mcnemar_results_phys}.

For Author 0, the general direction is that the baseline model performance is better, although this is notably due to a single increase in $n_{10} = 1$ for four of five seeds compared to $n_{01} = 0$ across all seeds. Simply put, there is a single item that was correctly predicted by the baseline model but incorrectly predicted by the fine-tuned model, while there are no items for corrected baseline errors (i.e., items incorrectly predicted by baseline but correctly predicted by fine-tuned model). This also reflects the trend in average ExpRate, which settled at fine-tune size of 20 on a $1.73\%$ decrease compared to base model ExpRate. This suggests that due to the already high ExpRate of the base model on the standard handwriting of Author 0, with 91.30\% exact matches, it became more difficult for the model in the fine-tuning process to identify further ``corrections" that would lead to more exact matches. As such, fine-tuning may have caused the model to slightly ``overcorrect" its prediction, though this only resulted in one single less exact match compared to the baseline, suggesting the effect of ``overcorrection" remained minimal for Author 0.

Meanwhile, for Author 1, the consistent direction is that the adapted model is better than the baseline model. Most individual $p$-values from the McNemar tests do not reach significance, with all $p$-values for rushed handwriting falling at $p > 0.10$. Item-level permutation test $p$-value suggests directional consistency for standard handwriting (at $p < 0.05$) but less so for rushed handwriting. Nonetheless, this does not indicate absence of the improvement effect for rushed handwriting, as both conditions still show a similar number of corrected baseline errors on average (with mean $n_{01} = 5.6$ for standard and mean $n_{01} = 5.0$ for rushed), but rushed handwriting shows roughly twice the rate of ``regressions" (mean $n_{10} = 1.4$ compared to $0.8$) caused by fine-tuning adaptation. This could point to adaptation on rushed handwriting still being beneficial but less ``stable".

\begin{table}[h]
\centering
\caption{ Results on McNemar's test performed on physical handwriting-adapted models across 5 seeds. Results on item-level permutation test are below each author table. }
\label{tab:mcnemar_results_phys}

\begin{subtable}[t]{\linewidth}
\centering
\begin{tabularx}{\linewidth}{c|c c c X}
\toprule
\textbf{Seed} & $n_{01}$ & $n_{10}$ & $p$-value & \textbf{Direction} \\
\midrule
(1) & 0 & 1 & 1.0000 & baseline better \\
\hline
(2) & 0 & 1 & 1.0000 & baseline better \\
\hline
(3) & 0 & 1 & 1.0000 & baseline better \\
\hline
(4) & 0 & 1 & 1.0000 & baseline better \\
\hline
(5) & 0 & 0 & 1.0000 & tied \\
\bottomrule
Permutation test & 0 & 4 & 0.5002 & \\
\end{tabularx}
\caption{Author 0 - standard handwriting}
\end{subtable}

\begin{subtable}[t]{\linewidth}
\centering
\begin{tabularx}{\linewidth}{c|c c c X}
\toprule
\textbf{Seed} & $n_{01}$ & $n_{10}$ & $p$-value & \textbf{Direction} \\
\midrule
(1) & 5 & 0 & 0.0625 & adapted better \\
\hline
(2) & 7 & 1 & 0.0703 & adapted better \\
\hline
(3) & 6 & 1 & 0.1250 & adapted better \\
\hline
(4) & 5 & 1 & 0.2188 & adapted better \\
\hline
(5) & 5 & 1 & 0.2188 & adapted better \\
\bottomrule
Permutation test & 28 & 4 & 0.0407 & \\
\end{tabularx}
\caption{Author 1 - standard handwriting}
\end{subtable}

\begin{subtable}[t]{\linewidth}
\centering
\begin{tabularx}{\linewidth}{c|c c c X}
\toprule
\textbf{Seed} & $n_{01}$ & $n_{10}$ & $p$-value & \textbf{Direction} \\
\midrule
(1) & 5 & 1 & 0.2188 & adapted better \\
\hline
(2) & 4 & 1 & 0.3750 & adapted better \\
\hline
(3) & 5 & 1 & 0.2188 & adapted better \\
\hline
(4) & 6 & 2 & 0.2891 & adapted better \\
\hline
(5) & 5 & 2 & 0.4531 & adapted better \\
\bottomrule
Permutation test & 25 & 7 & 0.1793 & \\
\end{tabularx}
\caption{Author 1 - rushed handwriting}
\end{subtable}

\end{table}

\section{Limitations}
We perform all custom-author experiments on only two authors. More authors and more samples per author would improve the statistical strength of observations regarding writer adaptation for HMER.

\section{Conclusion and future work}
We characterize the domain gap between the digital training images and the physical images used during inference. Results suggest an increase in ExpRate and decrease in CER for digital handwriting. However, results are more varied for physical handwriting, with the model struggling for handwriting articles of one author, where the base model can already evaluate with very high ExpRate, although writer-adaptation may have some resistance to overcorrection, as this only resulted in a 1.73\% decrease in ExpRate compared to the base model. For the physical handwriting of the other author, adaptation shows general increase in ExpRate and decrease in CER for both standard and rushed handwriting. As for attention and parameter adaptation of the MFH-CoMER, these are found to be largely writer-dependent. We recommend testing writer-adaptive MFH-CoMer using more types of fine-tuning and adaptation methods, as well as deeper integration into HMER architectures. The focus of this study is largely to determine potential feasibility of writer adaptation for the HMER task, and as such, the fine-tuning method and model are structurally simple. We also recommend testing with more authors and more samples for stronger results.

\bibliographystyle{IEEEtran}
\bibliography{refs}

\end{document}